\documentclass[conference]{IEEEtran}
\IEEEoverridecommandlockouts
\usepackage{cite}
\usepackage{amsmath,amssymb,amsfonts}
\usepackage{algorithm}
\usepackage{algpseudocode}
\usepackage{graphicx}
\usepackage{textcomp}
\usepackage{xcolor}
\usepackage{subfigure}
\def\BibTeX{{\rm B\kern-.05em{\sc i\kern-.025em b}\kern-.08em
    T\kern-.1667em\lower.7ex\hbox{E}\kern-.125emX}}
\begin{document}

\title{Combining Federated and Active Learning for Communication-efficient Distributed Failure Prediction in Aeronautics}

\author{\IEEEauthorblockN{Nicolas Aussel\IEEEauthorrefmark{1}\IEEEauthorrefmark{2},
Sophie Chabridon\IEEEauthorrefmark{2},
Yohan Petetin\IEEEauthorrefmark{2}}
\IEEEauthorblockA{\IEEEauthorrefmark{1}formerly: Zodiac Inflight Innovations, Wessling, Germany}
\IEEEauthorblockA{\IEEEauthorrefmark{2}SAMOVAR, T\'{e}l\'{e}com SudParis, Institut Polytechnique de Paris, France}}

\maketitle

\begin{abstract}
Machine Learning has proven useful in the recent years as a way to achieve failure prediction for industrial systems. However, the high computational resources necessary to run learning algorithms are an obstacle to its widespread application. The sub-field of Distributed Learning offers a solution to this problem by enabling the use of remote resources but at the expense of introducing communication costs in the application that are not always acceptable. In this paper, we propose a distributed learning approach able to optimize the use of computational and communication resources to achieve excellent learning model performances through a centralized architecture. To achieve this, we present a new centralized distributed learning algorithm that relies on the learning paradigms of Active Learning and Federated Learning to offer a communication-efficient method that offers guarantees of model precision on both the clients and the central server. We evaluate this method on a public benchmark and show that its performances in terms of precision are very close to state-of-the-art performance level of non-distributed learning despite additional constraints.
\end{abstract}

\begin{IEEEkeywords}
Distributed Learning, Failure Prediction, Active Learning, Federated Learning, Machine Learning
\end{IEEEkeywords}

\section{Introduction}
\label{introduction}

\subsection{General problem}

In the recent years, the efficiency of Machine Learning for automated processing of large volumes of data has been widely demonstrated. This has been of particular interest for industrial applications that  commonly generate large datasets. One such application is failure prediction as the ability to anticipate on a system failure can lead to improvement in its reliability and to more optimized maintenance processes. However, a known limitation for the use of Machine Learning is its high computational costs relative to more traditional threshold-based system monitoring. In some application contexts, a simple solution to this can be found by either deploying additional resources at the location where they are needed or by offloading the data to a remote location following the cloud computing paradigm. There are however situations where neither of those options are realistic.

If we consider the application case of vehicular systems in general and, more specifically, aeronautical systems, they would greatly benefit from increased reliability and optimized maintenance processes and the latest generations of aircraft do have both on-board computational capabilities and data links with the ground. However, aeronautical systems generate too much data to be able to handle failure prediction on the aircraft for lack of computational resources and deploying additional on-board resources is very complex and costly. Moreover, the data links available between the aircraft and the ground are too expensive and not fast enough to transfer all the data. Therefore, we cannot directly apply a cloud computing paradigm which would consist in offloading all the data to a remote location such as a data centre with sufficient computational resources and collecting the results.

The solution for this would be to distribute the workload between the aircraft and a ground station in a client/server architecture in order to use their different capabilities in the most efficient way. In this situation, we can identify requirements that the solution will need to demonstrate to function in a distributed manner. Firstly, it will need to be able to classify data from normal readings, i.e. situations where there are no signs of imminent failures, using a minimal amount of computation power. We will refer to this requirement as $R_1$. This is key to be able to process the large volume of data generated by the aeronautical systems without overloading the data link. Secondly, the solution for failure prediction will need a mechanism to identify interesting data, i.e. data that could be indicative of an imminent failure with regards to a pre-defined threshold, and transfer this data from the aircraft to a central server on the ground without spending too much communication budget. We will refer to it as $R_2$. This is a direct consequence of the observation that the amount of computational power on board is too limited to give an accurate prediction of failure risks. Thirdly, we need a way to train a prediction model on the ground without accessing the data from the aircraft. We will refer to it as $R_3$. This is a direct consequence of the previous two requirements as only a small amount data can be sent to the central server. But, as accurate predictions are expected from it, we need a mechanism to enable centralized learning without accessing the raw data.

Different solutions already exist for each of the requirements we have identified taken individually. The main concern here is to figure out how to fulfil all three requirements simultaneously.

\subsection{Distributed Learning paradigms}

The field of Machine Learning dealing with distributed algorithms is Distributed Learning. There are two sub-fields in particular that provide interesting elements of answer for our application case, Active Learning~\cite{cohn1996active} and Federated Learning~\cite{mcmahan2016communication}. They have been designed with slightly different use cases in mind which obviously leads to slightly different requirements but that nevertheless matches our use case quite well.

The Active Learning paradigm offers a way to model the envisioned relationship between the aircraft and the central server on the ground. Its premise is that the model being trained has access to a third-party source of information, called an oracle, to reveal labels when necessary and its goal is to maximize the performance of the trained model under a request budget. We can immediately draw a parallel with the requirement $R_2$ of the solution we identified, identify uncertain samples and transmit them under a communication budget. With the aircraft playing the role of the active learning model in training and the central server playing the role of the oracle albeit an imperfect one (an unusual situation but already studied, for example in~\cite{miller2014adversarial}), the active learning paradigm offers an approach to select the hard-to-classify data and balance the amount of requests with the communication budget. Regarding requirement $R_1$, enabling computation-efficient classification of normal samples, however, further investigations are required as a number of works in Active Learning favour a greedy approach incompatible with the idea that normal readings need to be processed quickly.

The Federated Learning paradigm is directly related to requirement $R_3$, i.e. enabling centralized learning without accessing raw data. The communication budget and the computational resources of the hosts are also concerns that are considered in this approach. However, it is purely driven by the central server and does not provide any mechanism for the clients to assess their performance and request clarifications from the central server.

The goal of this contribution is to propose a new approach that fulfils all three requirements by combining Active Learning and Federated Learning. The performances are evaluated on a public benchmark and compared to existing solutions.

The contribution is organized as follows. First, we study existing works in Active Learning and Federated Learning in section \ref{related}. Next, we detail our solution in section \ref{algorithm}. We present our results and interpret them in section \ref{results} and section \ref{conclusion} concludes the contribution.


\section{Related Work}
\label{related}

In this section, we review state-of-the-art works related to our problem focussing on the applicability of the techniques reviewed with regards to the three identified requirements.

On the matter of distributed learning publications that are of particular interest to us we have to mention~\cite{valerio2017communication}. In this article, an Internet of Things (IoT) distributed learning framework for classification is presented and evaluated on network overhead and model accuracy. The authors present a solution called Hypothesis Transfer Learning based on an implementation of Greedy Transfer Learning found in~\cite{10.1007/978-3-319-23231-7_1}. In this decentralized approach, every host computes a local model based on the data accessible to it, sends its model to all its neighbours in a synchronization phase and collects their models. A second round of training is done, using Greedy Transfer Learning from neighbour models to improve the local model. The synchronization phase is then repeated until the host receives the updated local model of every host that has contributed to the distributed learning. Finally, the models are aggregated in an ensemble model using majority voting, that is to say the result of the classification is the most frequent prediction among the aggregated models. This work considers several hypothesis that are relevant for aeronautical systems; namely, it avoids data transfer between hosts, provides an explicit way to trade-off network overhead and models accuracy by tuning how many nodes take part to the Greedy Transfer Learning training and it considers the case of class imbalance (though with a much less drastic ratio). The most significant difference with the prediction in aeronautical systems, beside the order of magnitude change in class imbalance ratio, is that it is a decentralized algorithm with communication between hosts, which is not a realistic hypothesis for an aircraft. Another problem that it does not address is the fact that the performance of individual models is not controlled meaning that it is not lower bounded and that there are no mechanisms to correct a potential drift of a local model. It makes sense in the situation considered in~\cite{valerio2017communication} but it is a requirement that needs to be addressed in the aeronautical use case.

Note that for aeronautics applications, we would like to use Random Forest (RF) because of its resilience to noise and class imbalance as seen in~\cite{AusselJGPFC17} and that it can be proved reliable in a range of safety scenarios. An issue remains in that RF is easy to distribute but there is no guarantee on the performance of individual trees.

\subsection{Federated Learning}

The terminology of Federated Learning was coined recently in~\cite{mcmahan2016communication}. It describes a Distributed Learning situation where the following observations apply:
\begin{itemize}
\item There is a massive number of clients with limited computational resources each holding a unique fraction of the dataset. Their availability is subject to changes. This matches well the aeronautical use case where individual aircraft can be seen as clients. Their availability is not perfect and depends on the satellite connection.
\item The data are not independent and identically distributed between clients. In the original use case because it is assumed that different users will interact differently with their mobile device and type differently. In the aeronautical use case, this holds true as different aircraft can face different flight conditions depending on their route and the way their systems are used by the crew and the passengers.
\item Communications are allowed between the clients and a central server but they are limited by a communication budget and, for privacy concerns, no exchange of raw data is allowed. In the aeronautical use case, the privacy requirement between the client and the server is relaxed but the communication constraint is certainly relevant.
\end{itemize}
The original solution proposed is to initialize randomly a deep learning model at the server level and to gather incremental Stochastic Gradient Descent (SGD) updates from the clients while exchanging only the model parameters with them.

Since the publication of the original article, several contributions have followed and improved different aspects of this method. More specifically,~\cite{konevcny2016federated} proposes a way to improve the communication efficiency and~\cite{DBLP:journals/corr/BonawitzIKMMPRS16} and~\cite{geyer2017differentially} offer new insights as how to better guarantee that the privacy concerns are respected.

The Federated Learning approach is of high interest for us as it describes a framework for massively distributed and communication-efficient learning in a context very close to ours. In particular, Federated Learning would be a great answer to our  requirement $R_3$, learning in a centralized model without downloading data from the clients. There are however limitations. We have so far identified RF as our most promising choice for a base learning model. Also, Federated Learning is developed for Deep Neural Networks (DNNs) and uses incremental SGD updates from the clients. However, RF does not use SGD so we cannot apply Federated Learning directly. Also, Federated Learning does not offer any solution for our second identified requirement $R_2$, identify and uncertain samples and transmit them under a communication budget. For that, we turn to Active Learning.

\subsection{Active Learning}

Active Learning is a sub-field of semi-supervised learning where a third-party, called an oracle, can provide missing labels on request. It is traditionally applied in situations where data labelling requires the intervention of a human operator that is both costly and slow. The goal is then to learn the most accurate model possible under a given request budget. The general principle of Active Learning is interesting for our problem because it allows us to dynamically balance uncertainty on client-side and communication budget which would offer us a solution to fulfil the requirement $R_2$, identify uncertain samples and transmit them under a communication budget. 

A summary of the principles of Active Learning from a statistical perspective can be found in~\cite{cohn1996active}.The most common approach for concrete applications is called pool-based active learning and described in~\cite{10.1007/978-1-4471-2099-5_1}. The assumption in this approach is that there is a large pool of unlabelled data and a smaller pool of labelled data available. The labelled data is used to bootstrap a tentative model and a metrics to measure the amount of information gain one can expect from a sample is defined. Some examples of metrics can be a distance metrics between the sample and the decision boundary of the model, the density of labelled data in the vicinity of the sample or the difference in the learned model with regards to the possible labels. Once the metrics is defined, the unlabelled samples that are expected to maximize the information gain are queried from the oracle then the tentative model is updated and the process is repeated until the expected information gain falls below a pre-defined threshold, meaning that no new label is expected to significantly change the model, or the query budget is spent.

However, pool-based approaches are not a very good fit for our aeronautical application case as they often use a greedy sampling approach meant to minimize the intervention of the oracle, often a human operator, with the limitation of being prohibitively expensive in terms of computation; a human operator is indeed still orders of magnitude slower and more expensive than a complex computation. Examples of this can be found in~\cite{Georgala:2014:SFA:2611040.2611059},~\cite{5206627} and~\cite{LI2007459} illustrating the versatility of pool-based Active Learning with three applications on very different use cases of respectively spam detection, image classification and network intrusion detection. In our case, however, we envision the oracle as another more powerful model faster and less expensive than a human operator and we assume that complex computations are not possible on the client-side.

There are other works on online active learning that adopt a stream-based approach. In that approach, the decision to request a label from the oracle must be done immediately and leads to more efficient decision from a computation point of view. Examples of this approach can be found in~\cite{bouguelia2013stream} for an application to document classification and in~\cite{smailovic2014stream} for an application to sentiment analysis.~\cite{de2017confidence} is an article of particular interest to this article as it proposes a new way to train decision trees using an active learning approach that focuses on minimizing the risk of selecting a sub-optimal split when a new leaf is added to the tree by computing confidence intervals.  To put it differently, when a new leaf is added to the tree, the comparison is made between the ideal decision tree that could have been made if the true distribution of the samples were known and the expected performance of the current tree. If the difference is then found to exceed a certain threshold, labels are requested to minimize the risk of selecting a sub-optimal split for the new leaf. The point that is the most relevant for this article is that, by moving away from a greedy approach, we can ensure that the computation cost remains manageable, making this algorithm suitable for an on-board implementation. Moreover, the notion of approaching the uncertainty in terms of bounded risks for the model to be sub-optimal instead of considering sample misclassification makes it possible to directly manage the trade-off between the communication budget and the quality of the model. The only remaining limitation is that this model is not distributed. Therefore, we still need to implement a mechanism to leverage remote data sources to create a centralized model in order to fulfil requirement $R_3$, enable centralized learning without accessing client raw data.

\section{Proposed Algorithm}
\label{algorithm}

\begin{figure*}[htbp]
	\centering
	\includegraphics[scale=0.5]{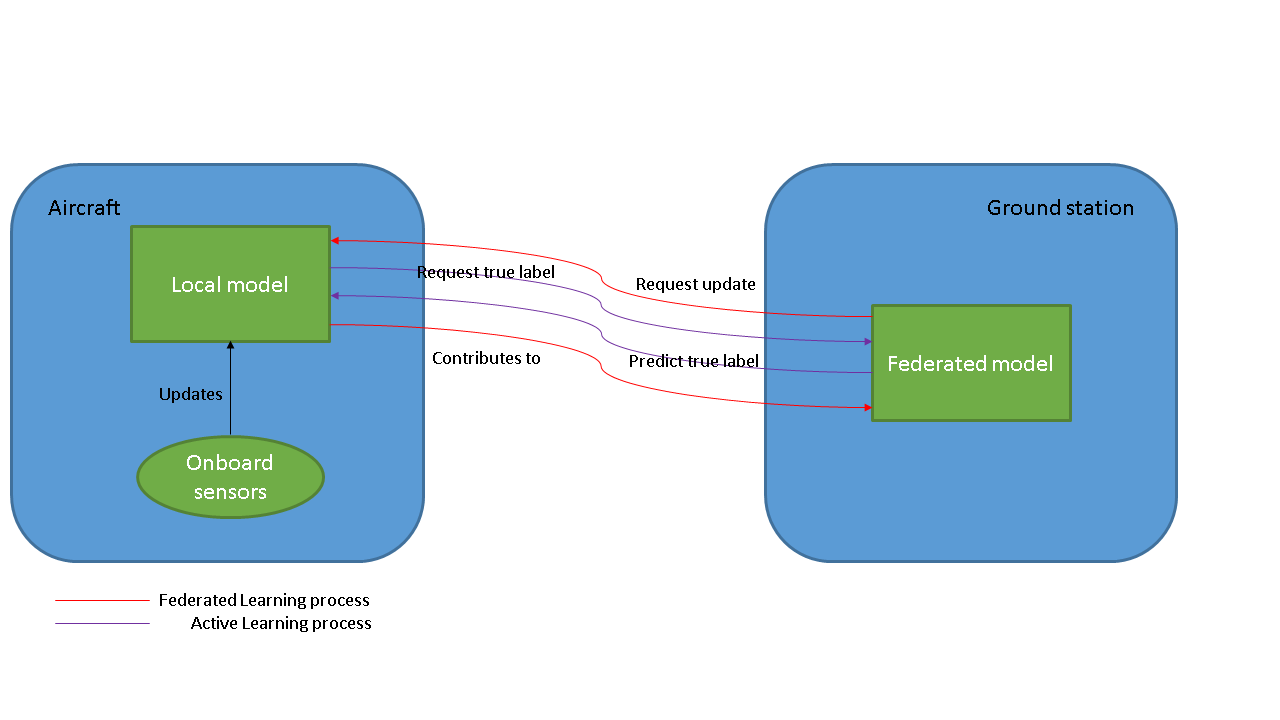}
	\caption{Illustration of Federated Active Learning}
	\label{fig:illustration_federated_active}
\end{figure*}

The approach we propose to combine Active Learning and Federated Learning is to use the confidence based active online Decision Tree (DT) from~\cite{de2017confidence} as a base model that is learned on the client side and sent to the server. The server aggregates the client DTs in an ensemble model similar to a RF to learn an accurate model without accessing the raw data. Finally, the individual DTs use the ensemble model from the server as an oracle for label requests. The figure~\ref{fig:illustration_federated_active} provides a high-level illustration of the approach envisioned.

To get more into details, each client has access to its local dataset and trains a confidence based active online DT locally.  Regularly, the central server sends an update request to the clients. The clients reply with their current DT. The server creates an ensemble model by collecting all the DTs. The ensemble model relies on majority voting to classify new samples, that is to say a new sample is run through each DT and its classification is the most frequent result among all DTs. A first parameter to control the communication budget of the approach will therefore be how often the central server requests updates from the clients.

From the clients perspective, they have at their disposal a stream of observations to train their local DT. As described in~\cite{de2017confidence}, whenever the uncertainty surrounding a new split in the DT exceeds a certain confidence threshold, a request is made to reveal the true label of the samples needed to decide on the right split and a part of the request budget is spent. Here the requests are sent to the server and the labels are decided using the ensemble model. There is a second parameter to control the communication budget of the approach in the request budget of each client.

There is a last point that needs to be clarified in the initialization of the model. Indeed, the necessary labels for the training of the DT are provided by the ensemble model. However, the ensemble model can only provide labels after it has received base models from the clients. In order to get around this constraint, we assume that some amount of labelled historical data is already available and can be used in the first round of update to initialize the DTs. This assumption is perfectly justified in the aeronautical use case where there is plenty of historical data available and the problem of label availability only concerns recent data where the true status of a system has not been clarified through a maintenance operation yet.

The proposed process is described in algorithm~\ref{Federated_active_alg}.

\algblock[]{Start}{End}
\begin{algorithm*}

	\caption{Federated Active Forest}
	\Comment{$F_{t}$ is the forest at round $t$, $T_{k,t}$ is the $k^{th}$ tree of $F_{t}$, $C$ is the set of clients, $C_{k}$ is the $k^{th}$ client of $C$, $n_{k,t}$ is the dataset gathered by the $k^{th}$ client between rounds $t$ and $t-1$, $B_{max}$ is the maximum request budget and $B_{k,t}$ is the request budget remaining for the $k^{th}$ client at round $t$}
		\begin{algorithmic}[1]
		\Start 
		\State $F_{0} \gets \Call{SeedForest}{C}$
		\For {each round $t=1,2,...$}
			\For{each client $C_k$ in $C$}
				\State $T_{k,t} \gets \Call{ClientUpdate}{(T_{k,t-1})}$
			\EndFor
			\State $F_{t} \gets \bigcup\limits_{\forall k}^{}(T_{k,t})$
		\EndFor
		\End		
		
		\Function{SeedForest}{$C$} \Comment{Server-side function}
			\For{each client $C_k$ in $C$}
				\State$T_{k,0} \gets \Call{SeedTree}$
			\EndFor
			\State $F_{0} \gets \bigcup\limits_{\forall k}^{}(T_{k,0})$
		\EndFunction
		
		\Function{ClientUpdate}{$T_{k,t-1}$} \Comment{Client-side function}
			\State $B_{updated} \gets B_{k,t-1} + \Call{UpdateBudget}{sizeof(n_{k,t})}$ \Comment{Request budget increased to match dataset growth}
			\State $T_{k,t}, B_{k_t} \gets \Call{ConfidenceDecisionTree}{T_{k,t-1}, n_{k,t} B_{updated}}$
			\State\Return($T_{k,t}$)
		\EndFunction	
		
		\Function{SeedTree}{} \Comment{Client-side function}
			\State $T_{k,0}, B_{k,0} \gets \Call{ConfidenceDecisionTree}{Root\_Tree, n_{k,0}, B_{max}}$ \Comment{$Root\_Tree$ is a tree limited to a root node, $n_{k,0}$ is assumed to be fully labelled}
			\State\Return($T_{k,0}$)
		\EndFunction	
		\end{algorithmic}
	\label{Federated_active_alg}
\end{algorithm*}

\section{Experimental Results}
\label{results}

\subsection{Experimental settings}

In order to enable a meaningful comparison with other works, we have selected for our experiments the MOA airlines dataset\footnote{https://moa.cms.waikato.ac.nz/datasets/}, one of the public benchmarks that was used in~\cite{de2017confidence}. Though it is related to the air transport industry, it is not strictly speaking a dataset from an aeronautical system but it presents the advantages of being public and enabling comparisons with other solutions. This dataset consists of over $500,000$ samples each with a label and seven features. Each sample corresponds to a real flight and the task is to predict if the flight has taken off on schedule or if it was delayed. The seven features available are summarized in Table~\ref{tab_MOA_features}. The imbalance ratio in this dataset is quite even at $0.80$ in favour of the flights that are not delayed. 
This means that some of the techniques such as Support Vector Machine (SVM) or Logistic Regression (LR)  could have acceptable performances on this particular dataset. However, since the target application for aeronautical systems would display a much higher level of class imbalance due to their reliability and since our proposed method is DT based, we do not consider them further.

\begin{table*}[htbp]
	\centering
	\begin{tabular}{@{}l|l|l@{}}
		Feature & Description & Format \\
		\hline
		Airline        & Unique identifier of the airline operating the flight & alphanumeric \\
		Flight numeric & Unique flight ID                                      & numeric      \\
		AirportFrom    & Origin airport unique IATA code                       & string       \\
		AirportTo      & Destination airport unique IATA code                  & string       \\
		DayOfWeek      & Day of the week                                       & numeric      \\
		Time           & Timestamp (scale not disclosed)                       & numeric      \\
		Length         & Length of the flight (min)                            & numeric      \\
	\end{tabular}
	\caption{Description of features in the MOA airlines dataset}
	\label{tab_MOA_features}
\end{table*}

To evaluate the performance of our algorithm, we report the precision, recall and F-score as in the previous contributions. Given the relatively balanced nature of this dataset and to provide as many elements of comparison as possible with other works done on this benchmark, we also report the accuracy. Defining the delayed flights as positives and the flights on schedule as negatives, as the labels in the dataset suggest us to do, and writing true positives, false positives, true negatives and false negatives respectively TP, FP, TN and FN, we define:
\begin{equation}
accuracy = \frac{TP + TN}{TP + TN + FP + FN}
\end{equation}
\begin{equation}
precision = \frac{TP}{TP + FP}
\end{equation}
\begin{equation}
recall = \frac{TP}{TP + FN}
\end{equation}
\begin{equation}
F-score = \frac{2 \times precision \times recall}{precision + recall} 
\end{equation}

In order to ensure the validity of our results, we apply a $2$-fold cross-validation by randomly splitting out the dataset into two sets, one for training and one for testing, measuring the performances a first time and repeating every measurement a second time after inverting the training and testing sets.
Regarding the parameters of this experiment, we examine more specifically the performance of our method with regards to the number of hosts among which the samples will be split, the maximum request budget available for the active learning process and the number of communication rounds for the federated learning process. When studying the variations of a parameter, others were kept at constant with $20$ communication rounds, $5$ clients and a request budget of $10\%$. The numbers of communication rounds and clients were chosen so that the constraints of distributed learning would be visible and the request budget was chosen so that comparisons with other methods are facilitated.

\subsection{Results and discussion}

\begin{figure}[htbp]
	\centering
	
	\begin{subfigure}[Accuracy]{
		\includegraphics[scale=0.5]{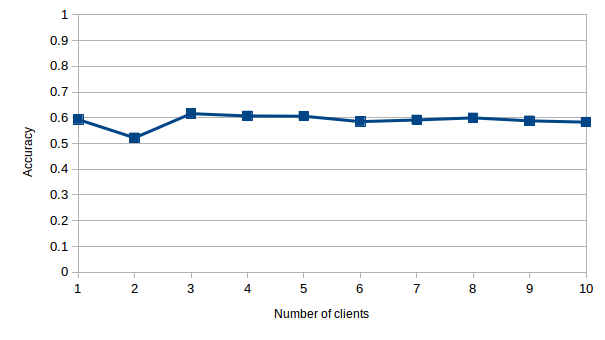}
		\label{federated_active_hosts_acc}
	}
	\end{subfigure}
	
	\begin{subfigure}[Precision]{
		\includegraphics[scale=0.5]{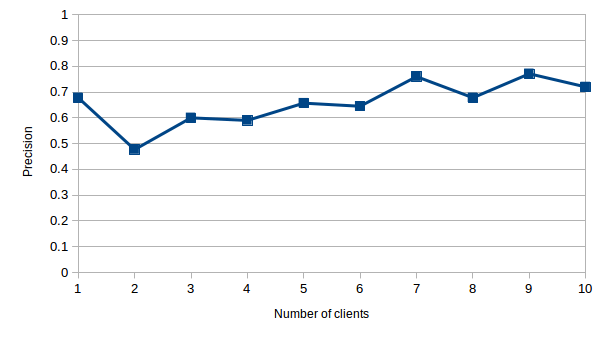}
		\label{federated_active_hosts_pr}
	}
	\end{subfigure}
	
	\begin{subfigure}[Recall]{
		\includegraphics[scale=0.5]{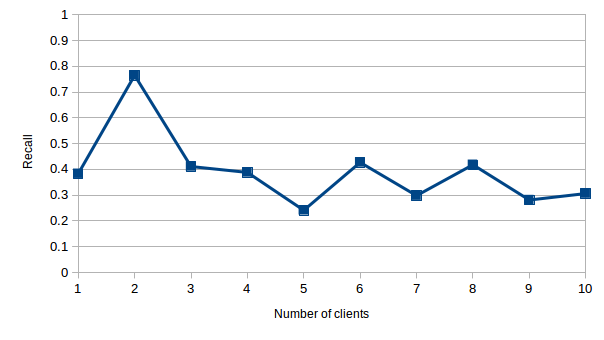}
		\label{federated_active_hosts_r}
	}
	\end{subfigure}
	
	\begin{subfigure}[F-score]{
		\includegraphics[scale=0.5]{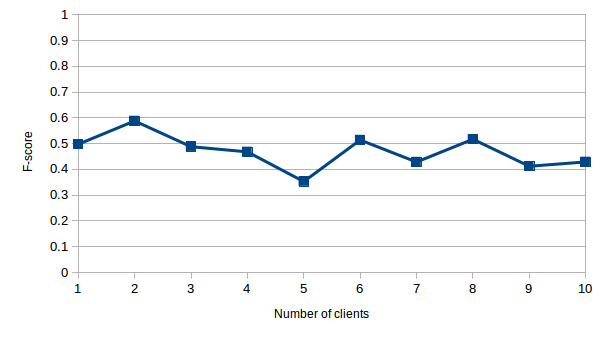}
		\label{federated_active_hosts_fs}
	}
	\end{subfigure}	
	\caption{Performance with regards to number of clients}
	\label{federated_active_hosts}
\end{figure}

\begin{figure}[htbp]
	\centering
	
	\begin{subfigure}[Accuracy]{
		\includegraphics[scale=0.5]{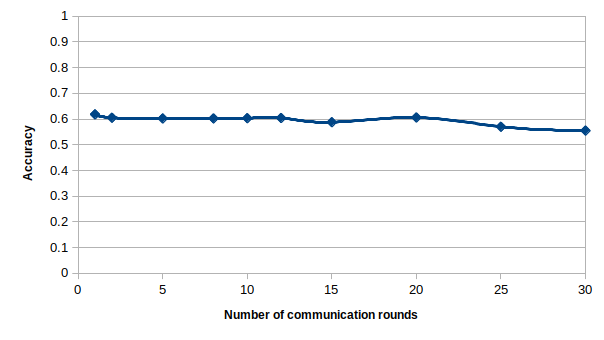}
		\label{federated_active_comm_acc}
	}
	\end{subfigure}
	
	\begin{subfigure}[Precision]{
		\includegraphics[scale=0.5]{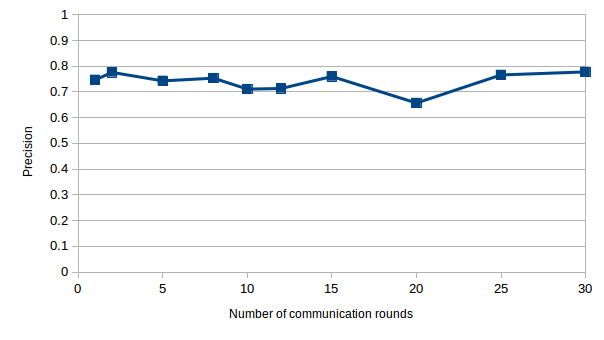}
		\label{federated_active_comm_pr}
	}
	\end{subfigure}
	
	\begin{subfigure}[Recall]{
		\includegraphics[scale=0.5]{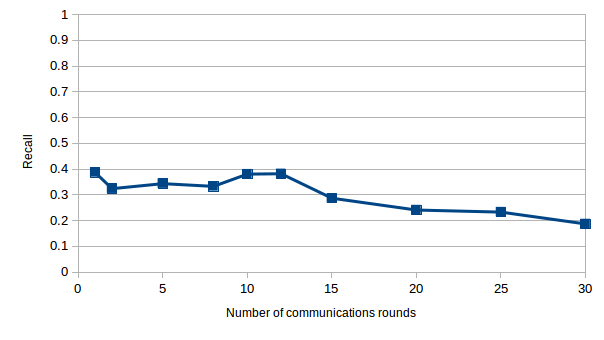}
		\label{federated_active_comm_r}
	}
	\end{subfigure}
	
	\begin{subfigure}[F-score]{
		\includegraphics[scale=0.5]{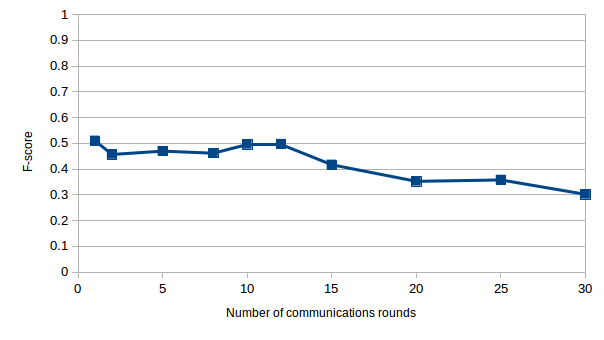}
		\label{federated_active_comm_fs}
	}
	\end{subfigure}	
	\caption{Performance with regards to the number of communication rounds}
	\label{federated_active_comm}
	
\end{figure}

\begin{figure}[htbp]
	\centering
	
	\begin{subfigure}[Accuracy]{
		\includegraphics[scale=0.5]{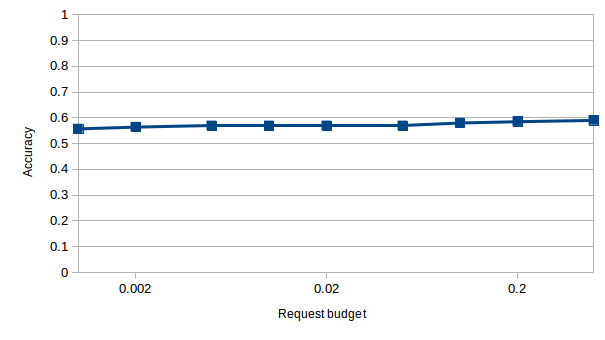}
		\label{federated_active_req_acc}
	}
	\end{subfigure}
	
	\begin{subfigure}[Precision]{
		\includegraphics[scale=0.5]{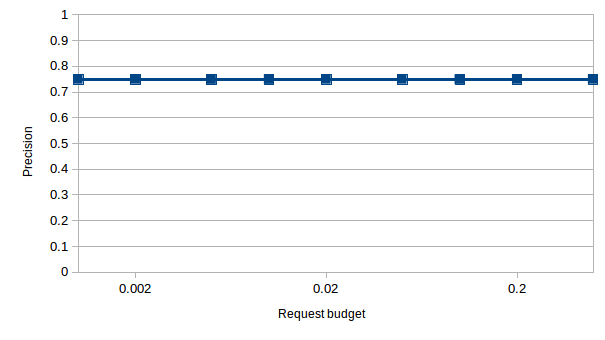}
		\label{federated_active_req_pr}
	}
	\end{subfigure}
	
	\begin{subfigure}[Recall]{
		\includegraphics[scale=0.5]{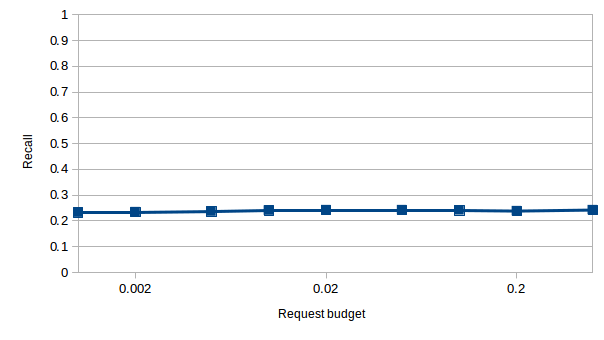}
		\label{federated_active_req_r}
	}
	\end{subfigure}
	
	\begin{subfigure}[F-score]{
		\includegraphics[scale=0.5]{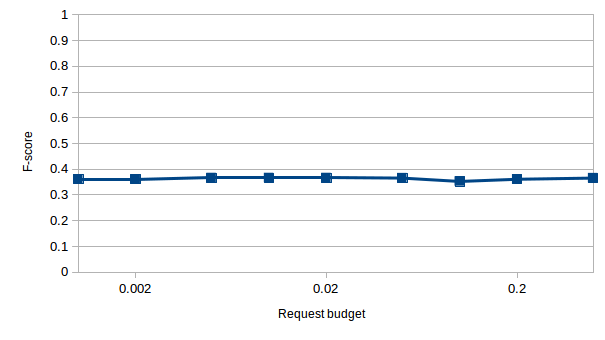}
		\label{federated_active_req_fs}
	}
	\end{subfigure}	
	\caption{Performance with regards to the request budget}
	\label{federated_active_req}
	
\end{figure}

\begin{figure}[htbp]
	\centering
	\includegraphics[scale=0.5]{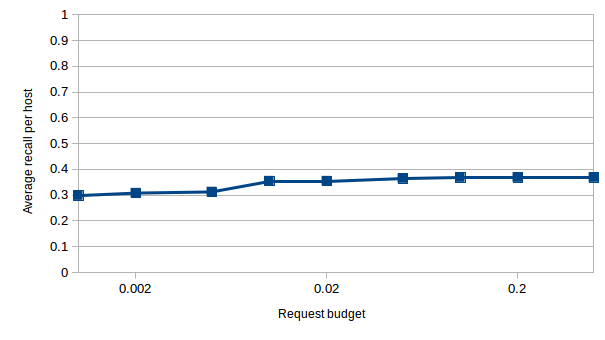}
	\caption{Average recall per client with regards to the budget request}
	\label{federated_active_req_per_host_acc}
\end{figure}

The results are grouped by variable parameter and displayed in Figure~\ref{federated_active_hosts} for the number of hosts, in Figure~\ref{federated_active_comm} for the number of communication rounds  and in Figure~\ref{federated_active_req} for the request budget. As additional information, Figure~\ref{federated_active_req_per_host_acc} provides insights on the average performances of individual clients for varying number request budgets.

In Figure~\ref{federated_active_hosts}, the first observation that we can make is that the results seem slightly unstable, especially considering that cross-validation has already been applied to them so one could expect smoother curves. In fact, the relative instability is a side effect of the majority voting employed in the ensemble model. When the number of clients is even, there is a limit case where each classification outcome gets the same number of votes. In such a case, our model was configured to favour the delay prediction, resulting in an increase in recall and a decrease in precision. This limit case is more frequent for small numbers of clients. Beside this, we can observe that the accuracy is stable at about $61\%$ which is an excellent level of performance comparable to state-of-the-art results in the non-distributed case with~\cite{de2017confidence} reporting about $62\%$ of accuracy with a similar request budget. This validates the potential of the approach we propose for communication-efficient distributed learning. For the other metrics, we can observe that they decrease slightly for a high number of clients which illustrates the fact that the distributed learning constraint does have an impact on the performance even if it is relatively small with the F-score decreasing from about $49.75\%$ with a single host to $41.3\%$ with $10$ hosts.

Concerning Figure~\ref{federated_active_comm}, the results are more straightforward. We observe that as the number of communication rounds increase, the model accuracy is quite stable but it hides the fact that precision is increasing at the expense of recall leading to an overall decreasing F-score from $51\%$ to $30\%$. The interpretation to this behaviour is that the ensemble model is being updated too frequently. The base DT model makes use of a grace period parameter which defines a minimum amount of samples that need to be collected before a request can be made. The default value used was $100$, the same as in the original publication. However, as the dataset is split between multiple clients and further down between multiple communication rounds, it is possible when the number of communication rounds is too high that between consecutive updates of the ensemble model, no updates of the local DTs have been made and consequently no change is made on the ensemble model leading to "skipped" communication rounds that eventually lead to lower the overall performances. We tested this interpretation by lowering the grace period to $30$ and observed that performances for a high number of communication rounds were increased.

Regarding Figure~\ref{federated_active_req}, it is worth noting that it is using a different scale than the other figures as the request budget is presented on a logarithmic scale. The reason for that is to highlight the extreme stability of the performances with a very slight increase of the accuracy for high request budget from $56\%$ to $59\%$. The interpretation for the very low values is largely due to the initialization step that relies on fully labelled data. Given $5$ clients, $20$ rounds and a $50/50$ training/test split, the amount of data used in the initialization represents $2.5\%$ of the total dataset so even for an extremely low request budget of $0.1\%$, it is necessary to label $2.51\%$ of the dataset. Still, given that typical range for request budget in similar studies such as~\cite{de2017confidence} starts at $10\%$ this is a very promising result, showing that our approach is efficient for extremely low request budgets thus fulfilling one of the requirements for applicability to aeronautical systems. Finally, we have included Figure~\ref{federated_active_req_per_host_acc} to give insight as to what changes in the model for different request budget. Indeed, if the performance of the ensemble model is very stable, it is hiding the fact that the performance of the local DTs. Selecting recall as the best illustration of this, we see that the average recall of individual DTs increases from about $30\%$ to $37\%$ when the request budget increases. This is significant as local DTs would act in a real-world implementation as back-up in situations where the data link between the aircraft and the ground would be unavailable, as such, being able to guarantee a certain level of performance of the local model is important.

Overall, it can be noted that the performances of the Federated Active Learning approach that we propose are very stable and very close to state-of-the-art level of performance for non-distributed learning despite the additional constraints.

\section{Conclusion}
\label{conclusion}

In this paper, we propose a new applied Machine Learning framework to enable real-time distributed learning for aeronautical systems. To do so, we identify three  requirements for a solution, namely $R_1$ the ability to classify normal samples with minimal computations, $R_2$ a mechanism to identify and transmit uncertain data under a communication budget and $R_3$ the ability to train a centralized model without accessing remote data. Then, we review existing solutions for each of the three requirements and propose a new method to fulfil all of them at the same time. Our method is a combination of Active Learning and Federated Learning. It relies on training confidence decision trees at client level and aggregating them at server level to create an ensemble model to act as an oracle. We provide a detailed performance analysis of our method on a public benchmark and compare our results to state-of-the-art classifiers.

Our approach achieves an accuracy of up to $61\%$ very close to state-of-the-art levels of performance of $62\%$ despite the additional constraint of Distributed Learning that implies that each client has to make decisions with partial information on the dataset and the constraint on limited computation power on the clients. Our method also achieves very consistent levels of performance with label budgets as low as $3\%$ thanks to the initialization step and the use of the ensemble model.

In general, with regards to the objective of this article of proposing a new communication-efficient and computationally cheap distributed learning framework suitable for the application case of real-time failure prediction for aeronautical systems, we have determined a new method that fulfils all three requirements we identified by combining existing works in an original way.

\bibliographystyle{IEEEtran}

\bibliography{biblio}

\end{document}